\documentclass[letterpaper, 10 pt, conference]{ieeeconf}  

\IEEEoverridecommandlockouts                              

\usepackage{graphics} 
\usepackage{epsfig} 
\usepackage{amsmath} 
\usepackage{amssymb}  
\usepackage{caption}

\usepackage{import}
\usepackage{multicol}
\usepackage{multirow}
\usepackage{comment}

\usepackage{microtype}

\usepackage{hyperref}
\hypersetup{
    colorlinks=true,
    urlcolor=blue,
}

\title{\LARGE \bf 
Demonstrate Once, Imitate Immediately (DOME):\\Learning Visual Servoing for One-Shot Imitation Learning}

\author{Eugene Valassakis$^{1}$, Georgios Papagiannis$^{1}$, Norman Di Palo$^{1}$, and Edward Johns$^{1}$
\thanks{This work was supported by the Royal Academy of Engineering
under the Research Fellowship scheme.}
\thanks{$^{1}$All authors are with The Robot Learning Lab at Imperial College London.
{\small Corresponding author: \tt eugene.valassakis15@imperial.ac.uk}}%
}

\begin{document}

\maketitle
\thispagestyle{empty}
\pagestyle{empty}


\begin{abstract}
We present DOME, a novel method for one-shot imitation learning, where a task can be learned from just a single demonstration and then be deployed immediately, without any further data collection or training. DOME does not require prior task or object knowledge, and can perform the task in novel object configurations and with distractors. At its core, DOME uses an image-conditioned object segmentation network followed by a learned visual servoing network, to move the robot's end-effector to the same relative pose to the object as during the demonstration, after which the task can be completed by replaying the demonstration's end-effector velocities. We show that DOME achieves near $100\%$ success rate on 7 real-world everyday tasks, and we perform several studies to thoroughly understand each individual component of DOME. Videos and supplementary material are available at: \url{https://www.robot-learning.uk/dome} .

\end{abstract}


\section{INTRODUCTION}

Imitation learning offers the potential for humans to teach robots how to perform a task, simply by providing a demonstration of that task. But many of today’s imitation learning methods are inefficient and difficult to deploy in practice.  Methods that learn tasks from scratch require significant data collection, such as multiple demonstrations \cite{zhang2018deep} or autonomous exploration \cite{schoettler2019deep,vecerik2017leveraging}. And methods that transfer knowledge from previous tasks \cite{finn2017one,duan2017one}, generalise poorly outside of that task family. In this paper, we ask the following question: \textbf{Can we design a method that can be deployed immediately following a single demonstration, without requiring any prior task or object knowledge?}

One of the reasons why most existing methods are so data inefficient and generalise poorly, is that they typically address imitation learning as an end-to-end policy learning problem. This requires explicit learning of the optimal action for all observations that might be experienced during deployment. Instead, in this paper we follow \cite{johns2021coarse} which proposes that if a robot's end-effector (EE) can be aligned with an object at the same relative pose as during the demonstration (the \textit{bottleneck pose}), then the demonstration's EE velocities can simply be replayed from the bottleneck onwards. But whereas \cite{johns2021coarse} required significant task-specific data collection following the demonstration, in this paper we ask if a new task could be executed immediately after observing the demonstration.

We approach the challenge of moving the EE to the bottleneck as a visual servoing problem, which consists of aligning a live image captured from a wrist-mounted camera, with the image captured at the bottleneck during the demonstration. However, with classical visual servoing come two limitations: (1) it requires prior object knowledge or strong geometric features~\cite{hutchinson2006visual}, which also, (2)  assume that the two images are observations of an identical scene~\cite{lowe2004distinctive}. To address the first we introduce a \textit{learned visual servoing network} which predicts the relative pose between two images, and to address the second we introduce an \textit{image-conditioned object segmentation network} which allows for visual servoing even under the presence of distractor objects. By training these two networks using simulated data over diverse objects and environment conditions, our method is able to perform visual servoing without prior knowledge of the object in the scene. Our full method, \textbf{D}emonstrate \textbf{O}nce, i\textbf{M}itate imm\textbf{E}diately (DOME), then solves a task by performing visual servoing until the bottleneck is reached, at which point the original demonstration's EE velocities are simply replayed.

\begin{figure}[t!]   
\centering    
\includegraphics[width=0.5\textwidth]{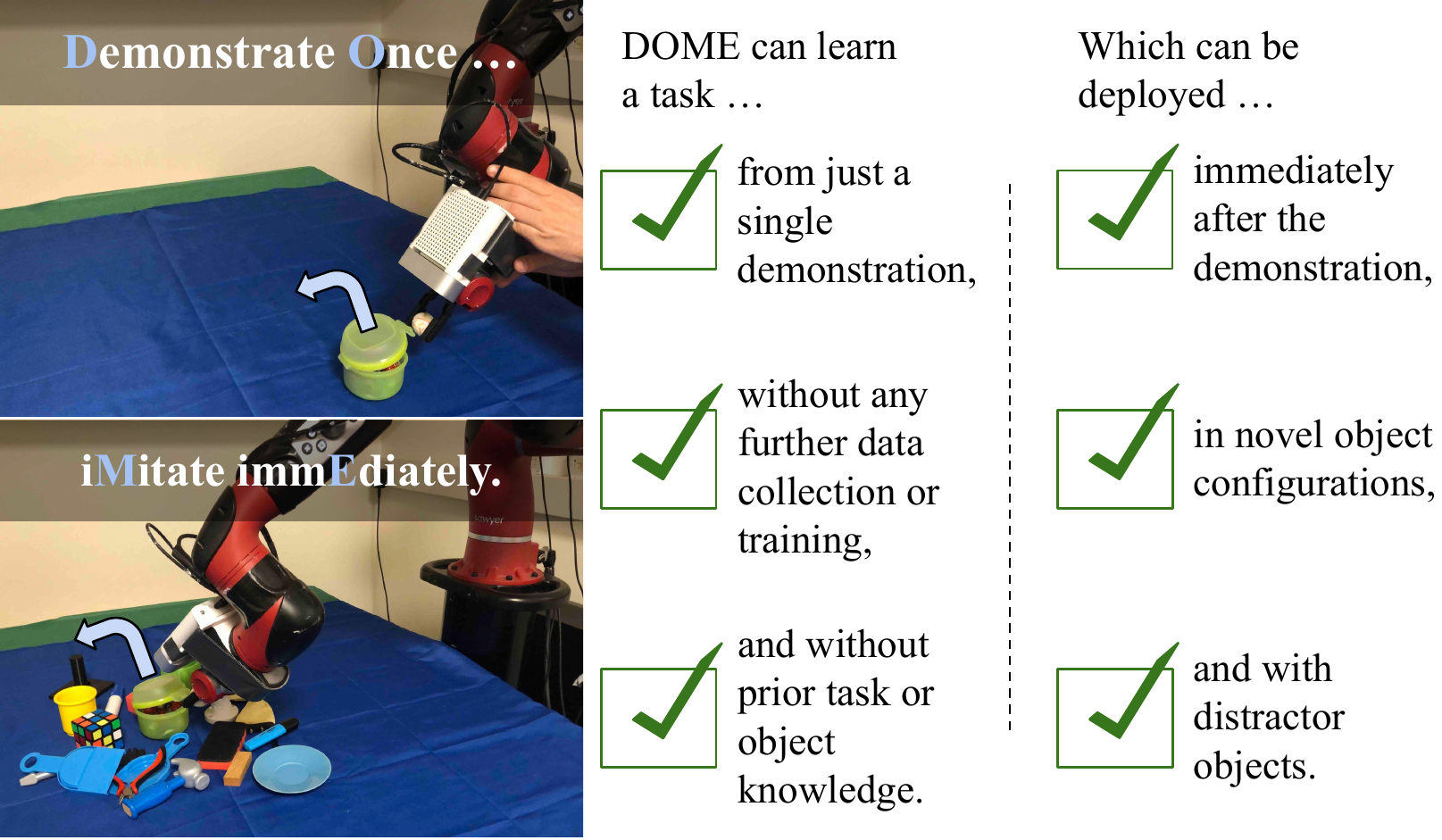}  
\caption{DOME can learn a task from just a single demonstration (in this example, opening the lid of a container).}
\label{fig:front_page}
\vspace{-0.6cm}
\end{figure}

To evaluate DOME, we designed a number of ablation studies which independently investigated the performance of each component of our overall framework. For our full framework evaluation, we then assessed its performance on 7 real-world everyday tasks. We found that DOME was able to perform these tasks immediately following just a single demonstration, and outperformed several baselines despite these baselines requiring further training after the demonstration. To the best of our knowledge, this is the first method that can learn a range of real-world tasks such as these from just a single demonstration, and then be deployed immediately, without requiring any further data collection or training after the demonstration. Fig.~\ref{fig:front_page} summarises the capabilities of DOME.

\section{Related Work}\label{sec:related-work}
\textbf{Imitation learning} (IL) methods aim to learn how to perform tasks from expert demonstrations~\cite{jang2022bc-z,zhang2018deep, osa2018algorithmic}. Within the field, behavioural cloning (BC) methods consist of learning to predict the expert's actions using supervised learning ~\cite{osa2018algorithmic,pemerleau1988alvinn, zhang2018deep}. Inverse reinforcement learning (IRL) methods \cite{ng2000algorithms,ziebart2008maximum} imitate an expert by inferring a reward function based on the provided demonstrations, which is then used to train a policy using standard reinforcement learning (RL). Other methods combining RL with demonstrations include \cite{vecerik2017leveraging, schoettler2019deep} which initialise the replay buffer of off-policy RL methods with demonstrations, or additionally augment their RL objective with a BC loss. Whilst able to learn tasks from scratch, these methods require a large number of demonstrations or expensive real-world interaction and human intervention (e.g. environment resetting) to be effective, making them impractical for real-world use. In our work, we address this issue by relying only on a \textit{single} demonstration and no further interaction to learn a task.

\textbf{One-shot IL} attempts to address this issue with methods that aim to solve a task from a single demonstration, without any further data collection. One set of approaches consist of meta-IL, which during a meta-learning phase train policies with the promise to quickly adapt to new tasks at test-time from only a single (or more generally, a few) demonstration(s)~\cite{finn2017one,duan2017one}. While the distribution of tasks used to train a meta-learner could theoretically incorporate a large variety of different tasks, current methods can mostly only deal with small variations within a task seen during meta-training \cite{finn2017one,duan2017one}. Another set of methods include approaches that design task-specific controllers (such as for pick and place~\cite{chai2019multi} or cloth folding~\cite{ganapathi2021learning}), and can imitate variations within this task. Closer to our work, FlowControl~\cite{argus2020flowcontrol} utilises optical flow for aligning to demonstration frames to complete a task. Unlike our method however, it relies on manually supplied object segmentations, and its core flow computation does not work well when the deployment frames differ significantly from the demonstration frames, or when there are background flows, which would occur with the addition of distractor objects. Finally, concurrently to our work, Wen et al.~\cite{wen2022you} developed a method for one-shot imitation with in-class object generalisation, but which assumes object class knowledge and has only been shown to work on placement and insertion-type tasks, whereas we show how our method can learn a range of task types.

\textbf{Coarse-to-fine robot manipulation} ~\cite{johns2021coarse,dipalo2021learning,valassakis2021coarse-to-fine,lee2020guided} models a task as two stages: an object reaching stage which can often be solved with kinematic planning, and an object interaction stage which requires more complex behaviour. It has been successfully applied to RL~\cite{lee2020guided}, sim-to-real control~\cite{valassakis2021coarse-to-fine}, and multi-stage IL~\cite{dipalo2021learning}. Following this principle, Coarse-to-Fine IL~\cite{johns2021coarse} is the closest method to our work. Like DOME, it solves IL by learning to reach a user-defined pose relative to the object of interest, followed by open-loop replay of the demonstration's EE velocities. However, it does not generalise to the presence of distractor objects, and requires lengthy self-supervised data collection and network training after the demonstration. DOME does not suffer from these drawbacks, and is able to immediately imitate a task even with the presence of distractor objects during deployment.

\begin{figure*}[!tbh]
\centering
  \includegraphics[width=1.\linewidth]{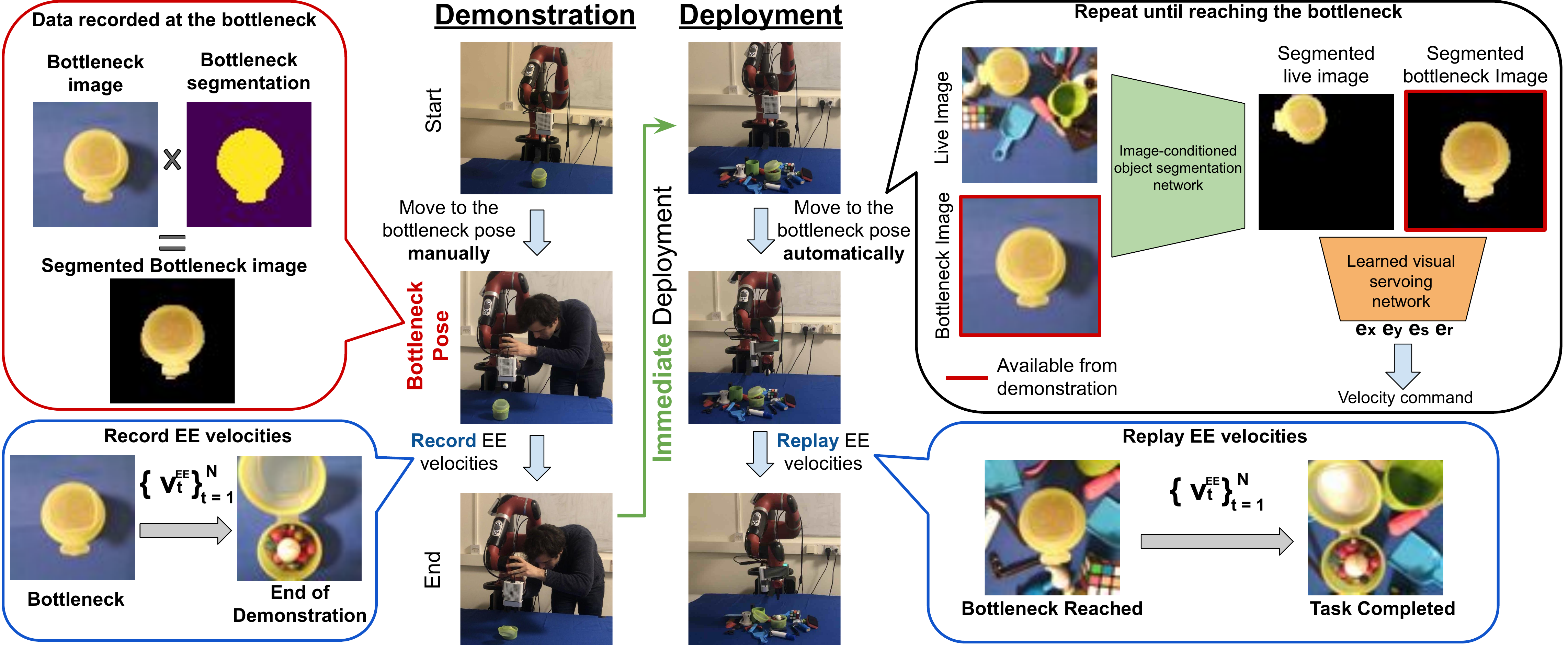}
  \caption{Overview of DOME, with the illustrated task involving opening the lid of the yellow box. EE: end-effector. \label{fig:method_diagram}}
  \vspace{-0.5cm}
\end{figure*}

\section{Methods}\label{sec:methods}

\subsection{Method Overview}

Inspired by the Coarse-to-Fine Imitation Learning framework~\cite{johns2021coarse}, our goal is to learn a controller that moves the EE to a particular pose relative to the object, called the \textit{bottleneck pose}, from which we can successfully complete a task simply by replaying the demonstration's EE velocities. 

An overview of our method is shown in Fig.~\ref{fig:method_diagram}. We use a wrist-mounted RGB camera (\textit{the camera}) that is fixed relative to the EE, such that the EE yaw axis and the camera's depth axis are parallel. First, the user gets to the \textit{bottleneck pose} and records an image from the camera which we call \textit{the bottleneck image}. The bottleneck pose is the starting point of the demonstration, and conceptually represents the pose the EE needs to reach before beginning any interaction with the object.  Second, the user performs the demonstration (see Section~\ref{subsec:demo}), and the system records the (linear and angular) EE velocities at each timestep. Third, deployment can start (see Section~\ref{subsec:deployment}).  The object can be repositioned at a different pose, and distractor objects can now be placed in the environment. During deployment, a learning-based visual servoing controller (see Section~\ref{subsec:deployment}) will move the EE to the bottleneck pose by aligning the current, \textit{live image}, it receives from the camera to the bottleneck image the user recorded at the start of the demonstration. When the two images are successfully aligned, then the EE has reached the bottleneck pose. At that point, demonstration replay can start. Since, at the bottleneck pose, the relative configuration between the object and the EE is the same as during the demonstration, replaying the demonstration EE velocities recorded in the EE frame will result in the same object interaction, thereby successfully completing the task.

\subsection{Demonstration} \label{subsec:demo}

In DOME, the aim of the demonstration is to record (1) the bottleneck image $I_{bot}$, (2) the \textit{bottleneck segmentation} $S_{bot}$, meaning a segmentation mask of the object in the bottleneck image, (3) the \textit{segmented bottleneck image} obtained through element-wise multiplication, $I^{seg}_{bot} = I_{bot} \times S_{bot}$, and (4) a sequence of EE velocities $\{v^{EE}_t\}_{t=1}^N$ that can complete the task from the bottleneck pose. This is illustrated in Fig.~\ref{fig:method_diagram}. To start a demonstration, the user first manually brings the robot's EE to the bottleneck pose. To do so the user can monitor the image from the camera, and stop when the object is fully visible and well centred in the image (see Fig.~\ref{fig:method_diagram}). At this point the user triggers the capture of the bottleneck image, and the bottleneck segmentation is computed by our simulation-trained image-conditioned object segmentation network (see Section~\ref{subsubsec:image-conditioned-seg}).  The user then provides the demonstration, and at each timestep the EE velocity in the EE frame is recorded.

\subsection{Deployment}\label{subsec:deployment}
Once the demonstration is finished, our method can be immediately deployed to imitate the task. At the beginning of deployment, we assume that the object of interest is at least partially visible from the camera.  As illustrated in Fig.~\ref{fig:method_diagram}, DOME consists of two separate controllers, a learning-based visual servoing controller that takes the EE from its initial pose to the bottleneck pose, and a demonstration replay controller that completes the task.

\subsubsection{\textbf{Learning-Based Visual Servoing Controller}} Our learning-based visual servoing controller has two main learned components. First, an image-conditioned object segmentation network that segments the 
object of interest in the \textit{live image} in order to obtain a \textit{live segmentation} of that object. To do so in the presence of distractors, it relies on the bottleneck image as conditioning information to be able to identify the corresponding object in the live image. Second, a learned visual servoing network that predicts velocities which move the EE towards the bottleneck pose. We discuss in detail our image-conditioned object segmentation network and learned visual servoing network in Section~\ref{subsec:learned-controllers}. 

 Our full control mechanism has the following 5 steps. At each control timestep $t$, we (1) input the live image $I_t$ and bottleneck image $I_{bot}$ to the image-conditioned object segmentation network $f$ to obtain the live segmentation $S_t= f(I_t,I_{bot})$, (2)  combine $S_t$ with $I_t$ to get a \textit{segmented live image} $I_t^{seg}= S_t \times I_t$, (3) input  $I_{bot}^{seg}$ from our demonstration data and $I_t^{seg}$ to our learned visual servoing network $f^{serv}$ to obtain the velocity in the camera frame $\mathbf{v^c_t}=f^{serv}(I_t^{seg}, I_{bot}^{seg})$  which will move the robot towards the bottleneck pose, (4) use kinematics to convert this velocity to joint velocity robot commands, and (5) repeat until the segmented live image and segmented bottleneck image are aligned, meaning that the bottleneck pose is reached, and which we detect by thresholding on the output of the visual servoing network $f^{serv}$.

\subsubsection{\textbf{Demonstration Replay Controller}} Similarly to~\cite{johns2021coarse}, demonstration replay consists of repeating the EE velocities recorded during demonstration. Formally, at each timestep $t$, we take $\mathbf{v^{EE}_t}$, the EE velocity in the EE frame recorded during demonstration, and map it using kinematics to joint velocity robot commands.

\subsection {Learned Components}\label{subsec:learned-controllers}
\subsubsection{\textbf{Image-Conditioned Object Segmentation Network}}\label{subsubsec:image-conditioned-seg} The ability to segment the object of interest during deployment allows our learned visual servoing network to focus only on that object, thereby naturally generalising to the presence of distractors. To do so, we build an image-conditioned object segmentation network that receives the  bottleneck and live images as inputs, and outputs a segmentation of the object of interest in the live image, allowing us to perform object detection and segmentation in a single forward pass. For simplicity, we will be referring to it as the \textit{segmentation network} from this point on-wards. We note that to obtain the bottleneck segmentation we also use this network, by simply using the bottleneck image twice in the input.  We investigate different possible network architectures for this problem in Section~\ref{subsec:cond-im-seg}. In this section, we present the architecture that we found worked best.

Our architecture is based on feature-wise linear modulation (FiLM)~\cite{perez2017film}, a method that was shown to be very effective for the adjacent problem of language conditioned object segmentation~\cite{perez2017film}. The idea behind FiLM is to pass the conditioning information through a neural network to obtain a set of parameters used to modulate the feature maps of a convolutional neural network (CNN), channel-wise. We refer the reader to~\cite{perez2017film} for the full details of FiLM. In our particular architecture, the live image is passed through a U-Net-like architecture in order to produce the object's segmentation map~\cite{unet}. Unlike U-Net however, the decoder feature maps of our network are modulated using FiLM. Finally, the modulating parameters for FiLM are the outputs of a CNN that uses the bottleneck image as its input. We train our network using supervised learning with ground truth segmentation masks available from simulation, and full training and architectural details can be found in our supplementary material.

\subsubsection{\textbf{Learned Visual Servoing Network}}\label{subsubsec:visual_servo_net} The other key learned component of DOME is the learned visual servoing network, which we will be referring to as the \textit{servoing network} for simplicity. It aims to output a velocity that when applied will move the live image towards the bottleneck image to align them.

In order to do so, a Siamese CNN~\cite{yu2019siamese} takes in the segmented live and bottleneck images and outputs (1) the pixel-space position difference between the object in the live and bottleneck images $\mathbf{e_{xy}} = [e_x,e_y]$, (2) its relative scale $e_s$, meaning the relative area occupied by the object between the bottleneck and live images, and used for control in the $z$ direction, and (3) the rotation $e_r$ of the camera between the live and bottleneck images.

In our experiments we assume that reaching the bottleneck by visual servoing is done in 4 Degrees of Freedom (DoF), so the bottleneck image is from a top-down view, $e_r$ is simply the rotation between the live and bottleneck images around the vertical axis, and the final velocity $\mathbf{v^{c}_t}$ is computed as follows:
\begin{equation}\label{eq:twist}
 \mathbf{v^{c}_t} =   \begin{bmatrix}
e_x*g_x/w \\
e_y*g_y/h \\
(clip(e_s, 0,2)-1)*g_s \\
0 \\
0 \\
{e_r*g_r}/\pi
\end{bmatrix},
\end{equation}

where $(w,h)$ are the height and width of the image, and $\mathbf{g} = \{g_x,g_y,g_s,g_r\}$ are $x,y$, scale and rotation gains applied to the predictions. Since $e_s$ is positive and theoretically unbounded, $clip$ restricts its maximum value to 2. Overall, assuming that the network outputs are correct, Equation~\ref{eq:twist} first normalises $\{e_{x}, e_{y},e_s,e_r\}$ to $[-1,1]$ individually, and then applies the gains such that $\mathbf{g}$  also represents the maximum linear and angular speed in each dimension, respectively. We note that since the gains here are set manually, this gives the opportunity to the user to tune the speed at which the learning-based visual servoing controller will get to the bottleneck pose. To verify whether this has any impact on performance, we have also conducted an ablation study with different values of these gains, which we present in Section~\ref{subsec:gain_ablation}.

Finally, we note that in practice we implemented the servoing network as three independent Siamese CNNs that predict $e_{xy}$, $e_s$ and $e_r$ from the same image inputs, which we found empirically to work better. For clarity, we refer to them as a single entity in this paper, and we refer the reader to our supplementary material for implementation and training details.

 \begin{figure}[!t]
\centering
 \hspace{-0.1cm}
  \includegraphics[width=1.\linewidth]{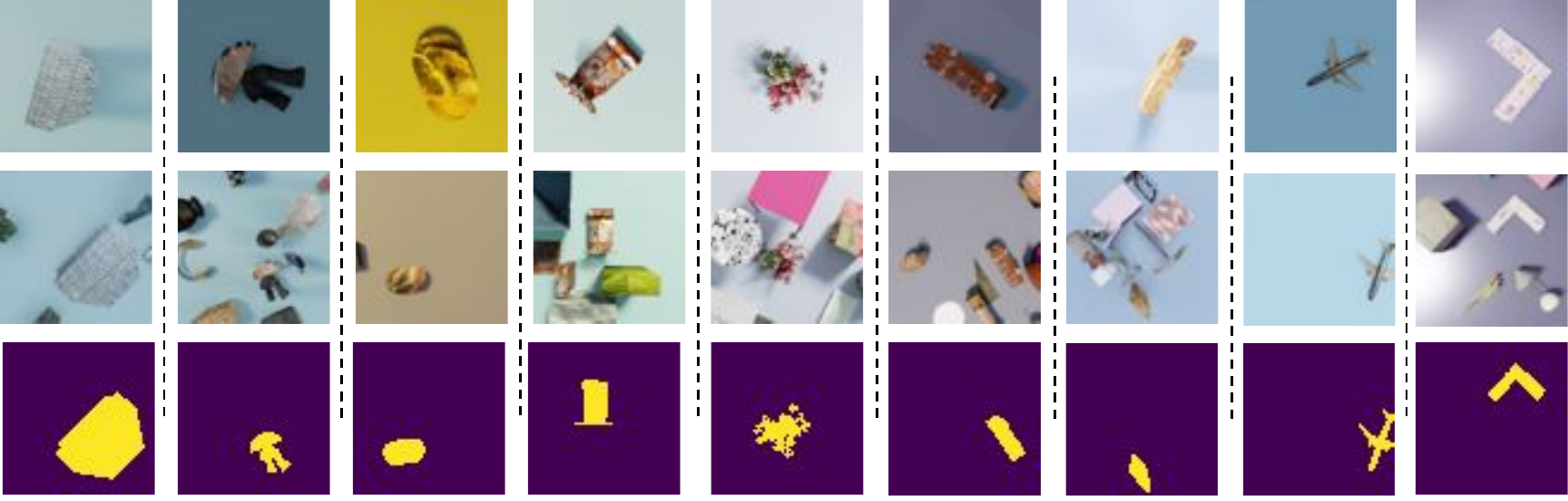}
  \caption{Simulated data examples. Top: Bottleneck image. Middle: Live image. Bottom: ground truth object segmentation. \label{fig:sim_data_examples}}
  \vspace{-0.2cm}
\end{figure}

\subsection{Training Data}\label{subsec:training-data}
We train all our models entirely in simulation using supervised learning. In order to generate the data, we use Blender~\cite{blender}, BlenderProc~\cite{denninger2019blenderproc},  and create a simulated environment with a virtual camera and random objects from Shapenet~\cite{shapenet2015}, and ModelNet40~\cite{wu20153d}, without considering what the downstream tasks may be. We first place a single object in the room, and take an image of the object to serve as the bottleneck image. We then place distractors in the scene, and take images from 4 new camera poses that serve as the live images. We also generate all necessary ground truth training labels during the process. In order to account for the reality gap~\cite{tobin2017domain}, we randomise various simulated environment conditions, such as (1) light parameters including the number, colour, location and strength of light sources, (2) the number of distractor objects present in the live images, and (3) object colour and texture parameters. For the object colour and texture parameters we note that we randomly either only change the base object model's surface parameters, such as colour or roughness, or additionally assign random low-fidelity images as textures to the surface of the objects. An ablation where we never assign images as textures can be found in Section~\ref{subsec:segm-quality-ablation}. In total, we generate 160,000 datapoints (i.e images and labels) for the segmentation network's training and 80,000 datapoints for the servoing network. Examples of our simulation data can be seen in Fig.~\ref{fig:sim_data_examples}, and details of our dataset generation pipeline can be found in our supplementary material.

 \begin{figure}[!t]
\centering
 \hspace{-0.1cm}
  \includegraphics[width=0.9\linewidth]{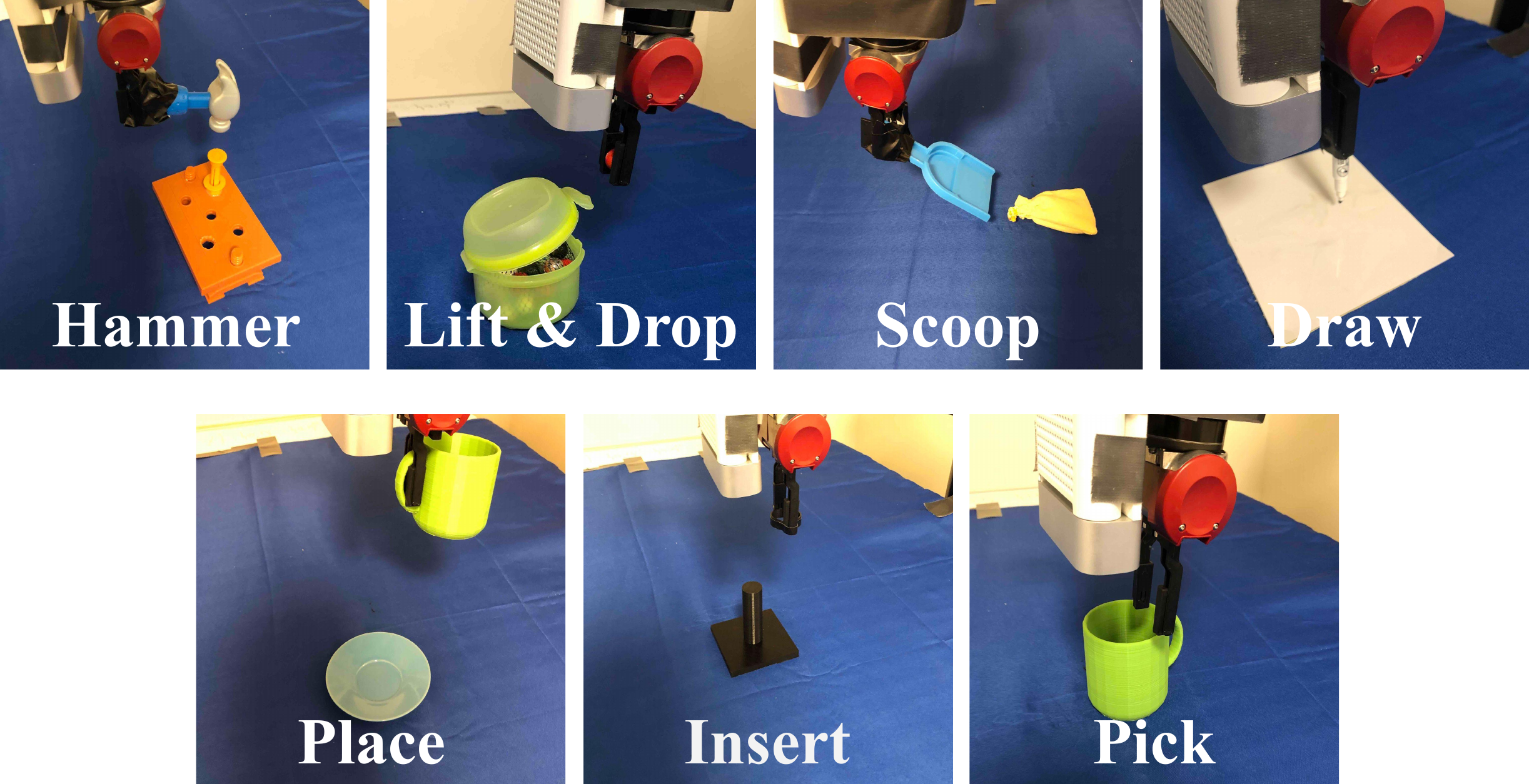}
  \caption{Illustration of our evaluation tasks. \label{fig:tasks}}
  \vspace{-0.4cm}
\end{figure}

\section{Experiments}

\subsection{Experimental Setup}\label{subsec:exp-setup}
For all our real-world manipulation experiments we used a 7 DoF Rethink Sawyer robot, to which we mounted a Microsoft Azure (that we used to generate RGB images, ignoring the depth) on the wrist. The images from the camera had $1280\times720$ resolution, that we cropped and resized to $64\times64$. We set the initial EE pose such that it lies approximately $45cm$ above the table. The task space (where we placed the objects) was defined such that the object would always be at least partially in view in the image at the EE's initial pose, giving an area of approximately $20cm\times15cm$. Supplementary material details and illustration videos are available at: \url{https://www.robot-learning.uk/dome} .

\subsection{Tasks}\label{subsec:tasks}
In order to evaluate our method, we chose 7 everyday manipulation tasks that are illustrated in Fig.~\ref{fig:tasks}. A \textit{Hammer} task where a plastic nail is knocked into a hole with a plastic hammer that is fixed to the EE. A \textit{Lift \& Drop} task where the lid on a plastic container is lifted and a pre-grasped ball is dropped inside. A \textit{Scoop} task where a small bag is scooped using a plastic shovel that is fixed to the EE. A \textit{Draw} task where a triangle is drawn on a plastic whiteboard sheet using a marker fixed to the EE. The sheet is placed on a  $\approx 5mm$ thick sponge to provide cushioning for safety, which in turn gives a maximum of $\approx 5mm$ error tolerance in the vertical direction for completing the task successfully. A \textit{Place} task where a pre-grasped cup is placed onto a small circular plate on the table.  An \textit{Insert} task where a circular ring fixed to the EE is inserted in a circular peg, with an $\approx 3.75mm$ error tolerance for successful insertion. And finally, a  \textit{Grasp} task where a plastic cup is grasped and lifted off the table.

\begin{table*}[t]

\small
\centering
\caption{Evaluating DOME against baselines by measuring the task success rate on each task. Best scores are in bold. \label{table:main_result}}
\resizebox{\linewidth}{!}{
\begin{tabular}{lccccc|ccccc}
                                            & \multicolumn{5}{|c|}{\textbf{Success Rate Without Distractors}}                                           & \multicolumn{5}{c}{\textbf{Success Rate With Distractors}}                                              \\ \hline
\multicolumn{1}{c|}{\textbf{Tasks}}         & \textbf{DOME (ours)} & \textbf{DOME H (ours)} & \textbf{CTF}   & \textbf{BC} & \textbf{RRL} & \textbf{DOME (ours)} & \textbf{DOME H  (ours)} & \textbf{CTF} & \textbf{BC} & \textbf{RRL} \\ \hline
\multicolumn{1}{l|}{\textbf{Hammer}}        & \textbf{100\%}       & \textbf{100\%}         & 73.3\%         & 0\%         & 0\%          & \textbf{100\%}       & \textbf{100\%}          & 60\%         & 0\%         & 0\%          \\
\multicolumn{1}{l|}{\textbf{Lift \& Drop}} & \textbf{100\%}       & \textbf{100\%}         & 6.7\%          & 0\%         & 0\%          & \textbf{93.3\%}      & \textbf{93.3\%}         & 0\%          & 0\%         & 0\%          \\
\multicolumn{1}{l|}{\textbf{Scoop}}         & 93.3\%               & 93.3\%                 & \textbf{100\%} & 0\%         & 0\%          & \textbf{93.3\%}      & 80.0\%                  & 20\%         & 0\%         & 0\%          \\
\multicolumn{1}{l|}{\textbf{Write}}         & \textbf{100\%}       & 93.3\%                 & \textbf{100\%} & 0\%         & 0\%          & \textbf{100\%}       & 80.0\%                  & 93.3\%       & 0\%         & 0\%          \\
\multicolumn{1}{l|}{\textbf{Insert}}        & \textbf{100\%}       & 93.3\%                 & 46.7\%         & 0\%         & 0\%          & 86.7\%               & \textbf{93.3\%}         & 6.7\%        & 0\%         & 0\%          \\
\multicolumn{1}{l|}{\textbf{Pick}}          & \textbf{100\%}       & \textbf{100\%}         & \textbf{100\%} & 0\%         & 0\%          & \textbf{100\%}       & \textbf{100\%}          & 13.3\%       & 0\%         & 0\%          \\
\multicolumn{1}{l|}{\textbf{Place}}         & \textbf{100\%}       & \textbf{100\%}         & \textbf{100\%} & 0\%         & 0\%          & \textbf{100\%}       & \textbf{100\%}          & 0\%          & 0\%         & 0\%         
\end{tabular}}
\vspace{-0.4cm}
\end{table*}

\begin{table}[t]
\small
\centering
\caption{Comparison of the mean real-world training overhead required for each method. Best scores are in bold. \label{table:training_time}}
\begin{tabular}{l|c|c|c}
                                                 & \multicolumn{1}{c}{\textbf{\begin{tabular}[c]{@{}c@{}}Training \\ time (mins)\end{tabular}}} & \multicolumn{1}{c}{\textbf{\begin{tabular}[c]{@{}c@{}}Training \\ samples\end{tabular}}} & \multicolumn{1}{c}{\textbf{\begin{tabular}[c]{@{}c@{}}Data collection\\  time (mins)\end{tabular}}} \\ \hline
\multicolumn{1}{l|}{\textbf{CTF}}                & 18.53                                                                                        & 12,723                                                                                & 40.0                               \\
\multicolumn{1}{l|}{\textbf{BC}}                 & 0.49                                                                                         & 288                                                                                & \textbf{0.0}             \\
\multicolumn{1}{l|}{\textbf{RRL}}                 & 4.67                                                                                            & 9,755                                                                                & 35.33         \\
\multicolumn{1}{l|}{\textbf{DOME H}}        & \textbf{0.0}                                                                                     &\textbf{0}                                                                          & \textbf{0.0}          \\
\multicolumn{1}{l|}{\textbf{DOME}} & \textbf{0.0}                                                                                         & \textbf{0}                                                                                    & \textbf{0.0}                                                                
\end{tabular}
\vspace{-0.5cm}
\end{table}

\subsection{Full Framework Evaluation}\label{subsec:overall-method-evaluation}

\subsubsection{\textbf{Baselines}}  To evaluate DOME we compared against 4 baselines.  For a fair test, we chose baselines which, like DOME, do not require prior task or object knowledge, and which only require a single demonstration.

First, a \textit{Behavioural Cloning} (BC)  baseline allowing us to test whether our tasks can be simply learned by supervised learning from a single demonstration. For this baseline we recorded a single demonstration completing the full task from the initial EE pose (as opposed to the bottleneck pose), storing EE velocities and wrist camera images during the process. We then trained a policy that maps wrist camera images to EE velocities on that demonstration, using the standard mean squared error loss, and deployed it for evaluation.

Second, a \textit{Residual Reinforcement Learning} (RRL) baseline to verify whether we can solve our tasks by allowing for environment exploration using RL. For this baseline, we first performed a single demonstration and trained a BC policy as described above. Then, inspired by~\cite{schoettler2019deep, johannink2019residual}, we used RL in order to learn corrective actions on top of this pre-trained BC policy.  Our reward consists of the L1 distance between the final image of the demonstration trajectory and the live image during exploration, similarly to~\cite{schoettler2019deep}. As our RL algorithm, we used TD3~\cite{fujimoto2018addressing}. Finally, we allow for a total of 40 minutes of environment interaction, which corresponds to environment exploration and training.

Third, a \textit{Coarse-to-Fine IL} (CTF) ~\cite{johns2021coarse} baseline, which is the method closest to our work. CTF uses self-supervised data collection in the real world to learn the transformation between the bottleneck pose and the EE's pose during deployment. We use the Prior Filtering and last-inch correction options presented in~\cite{johns2021coarse}, but instead of using a one-step prediction for the last-inch correction part, we use five. This is to account for the larger range in the angles of the object evaluation poses in our experiments, compared to~\cite{johns2021coarse}.  Finally, we allow for at total of 40 minutes of environment interaction, which corresponds to the self-supervised data collection.

Fourth, a \textit{DOME with hand-crafted visual servoing} (DOME H) baseline, to compare our servoing network that is learned to a hand-crafted alternative. As such, in DOME H, (1) $e_{xy}$ is obtained by taking the difference between the segmentation median pixel positions in the live and bottleneck segmentations, (2) $e_s$ is obtained by computing the ratio of the total segmented pixels in the bottleneck and live segmentations, and (3) $e_r$ is obtained using template matching for the two segmentations, by rotating the live image segmentation in increments of $1^{\circ}$, and selecting the rotation that gives the largest intersection over union (IOU) score with the bottleneck segmentation. We note that this hand-crafted controller is designed to work for these experiments with 4 DoF visual servoing for reaching the bottleneck, but, unlike our servoing network, does not have the potential to scale to the broader 6 DoF case.

\subsubsection{\textbf{Procedure}}\label{subsubsec:procedure} In order to evaluate these methods in the real-world, we predetermined 30 object poses in our task space, which we used for every method and every task. We obtained them by first uniformly sampling 30 positions in our task space and then sampling 30 rotations around the vertical axis, in the range $[-90^{\circ} ,90^{\circ}]$.

 When evaluating a method for a task,  we first performed a single demonstration and allowed any necessary data collection and network training to take place. Then, the method was deployed on each of the predetermined 30 object poses. For 15 of those object poses, we also added random distractor objects on the task space in order to evaluate how the methods respond to such a change between demonstration and deployment. We separately recorded successful task completion rates for the distractor and non-distractor cases. We present those results in Table~\ref{table:main_result}. For each method, we also recorded the number of \textbf{real-world} training samples,  network training time and the data collection time it took to train it. We present those results in Table~\ref{table:training_time}.  We note that since all methods have been provided with one demonstration, we did not take it into account when calculating the data collection time.  Finally, we note that for the RRL baseline, the data collection and network training are intertwined. As such, in order to obtain the training time, we accumulated the amount of time it took for the network back propagation. Then, to obtain the data collection time, we subtracted the training time from the pre-determined 40 minutes of total environment interaction time.

\subsubsection{\textbf{Results}}
From Tables~\ref{table:main_result} and~\ref{table:training_time}, we first see that our method shows great performance on all tasks, is robust to the presence of distractors, and does not require any further data collection or training time after the single demonstration. Particularly, we accredit the robustness to distractors to our segmentation network. It isolates the object of interest, thereby shielding the servoing network from being influenced by the presence of distractors. We also qualitatively note that failure cases are most often due to failures in the segmentation network, which then also induce errors in the servoing network outputs.

Second, as also shown in~\cite{johns2021coarse}, we see that CTF shows great performance in most tasks, although it uses a total of approximately an hour of data collection and training after the demonstration to achieve these results. Moreover, CTF is not robust to the presence of distractors, which is reasonable since it was never trained to do so.

Finally, we see that the RRL and BC baselines fail to complete any of the tasks. For BC, we believe this is because a single demonstration is not enough to learn a controller that is able to generalise. For RRL, we believe these tasks are too complex due to the difficulty of obtaining (and therefore the lack of) (1) an informative, useful, dense, reward, and (2) a well performing, pre-trained BC policy to guide exploration.

\subsection{Controller Gains Ablation}\label{subsec:gain_ablation}
\subsubsection{\textbf{Motivation}}
As we noted in Section~\ref{subsec:deployment}, with DOME the gains of the servoing network, and therefore overall speed of reaching the bottleneck, are set manually by the user. In this experiment we aimed to determine whether there is a trade-off between that speed and the performance of the controller, due for example to possible blurring in the captured images.

\subsubsection{\textbf{Procedure}} In order to determine this we evaluated DOME on the 15 object poses without distractor objects, as described in Section~\ref{subsubsec:procedure}.  We increased the gains of the servoing network by a factor of $\times3$, $\times5$, and $\times7$, which also increases the average speed for reaching the bottleneck by the same factor, and recorded the success rate on the 15 object poses on each task. We note that gains$\times7$ (1) corresponds to maximum speeds of $\{0.49m/s, 0.59m/s, 0.105m/s, 2.8rad/s\}$ in the $x,y,z$ and angular dimensions, and (2) it represented the setting where the robot could reach the bottleneck as fast as it is realistically safe.

\subsubsection{\textbf{Results}} In this experiment DOME achieved $100\%$ success rate across all tasks and gain values tested. As such, it became apparent that DOME is robust to changing the values for the servoing network gains. This is a desirable property, since it allows for the user to increase the overall speed of execution by increasing the speed of reaching the bottleneck.

\subsection{Segmentation Quality Dependence Evaluation}\label{subsec:segm-quality-ablation}
\subsubsection{\textbf{Motivation}}In order to test the robustness of our method to errors from our segmentation network, in this experiment we aim to answer the following question: How does the servoing network of DOME respond to varying quality of object segmentation?

\subsubsection{\textbf{Procedure}} To answer this question, we evaluated our servoing network in simulation. We performed a controlled experiment where we did not use the segmenetaion network. Instead, we used the ground truth segmentation masks, to which we added various degrees of noise before inputting them to our servoing network. 

In order to do so, we created a simulated version of our experimental procedure described in~\ref{subsec:overall-method-evaluation}, but with random objects selected from ModelNet40~\cite{wu20153d}, and only considering the task of reaching the bottleneck. We then defined the following set of perturbations to the segmentation masks used to create the inputs to our servoing network. \textit{Perfect} means we used the ground truth segmentation masks available from the simulator. \textit{Perlin} means we warped those masks with a Perlin noise vector field that remained constant throughout the trajectory.  \textit{Extra 1, 2, 3} means that we added 1, 2 or 3 extra segmentation artifacts to the live image (randomly selected regions added to the segmentation mask). Examples of images with these noise settings, as well as the Perlin noise parameters used are available on our supplementary material.

\begin{figure}[tb]   
\centering    
\includegraphics[width=0.45\textwidth]{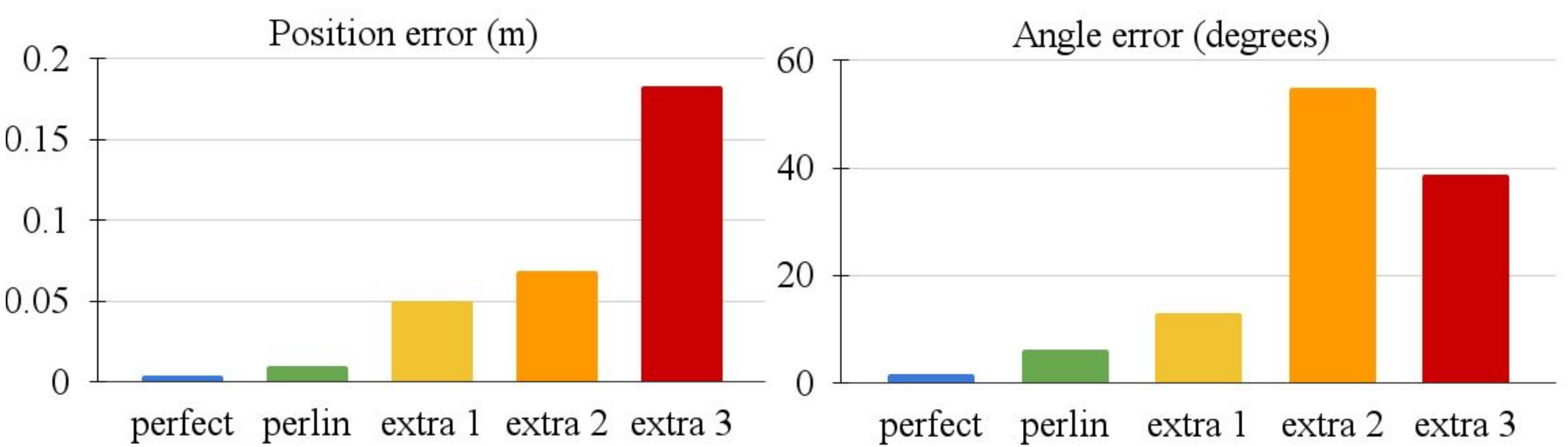}  
\caption{Servoing network performance for various levels of segmentation noise.}
\label{fig:seg_effect}
\vspace{-0.1cm}
\end{figure}

For each noise setting we deployed our servoing network for 50 trajectories, each with a different object to approach, and recorded the final position and angle errors. We report in Fig.~\ref{fig:seg_effect} the average position error and the median angle error across the 50 trajectories. We note that we report the median angle error because for symmetric objects a large angle error may appear even when the servoing network has produced a correct solution, and the median is less affected from such outliers.

\subsubsection{\textbf{Results}}  
As we can see from Fig.~\ref{fig:seg_effect}, the performance of the servoing network deteriorates significantly with increasing amounts of segmentation noise, especially when segmentation artifacts are introduced. The mean position error increases from a few millimetres to almost $20cm$, and the median angle error from below $2^{\circ}$ to more than $50^{\circ}$. This makes sense, since our servoing network is only trained with ground truth segmentation masks. We note that in practice how much visual servoing error will cause a task to fail is task-dependent. Finally, we also qualitatively noticed in our experimentation that a task is more likely to fail when the noise in the segmentation occurs in the bottleneck image (as opposed to the live image), since live image segmentation errors tend to average out over the course of a trajectory.

\begin{table}[t]
\small
\centering
\caption{ Perfomance of our image conditioned object segmentation models on the $160,000$ image dataset. Best scores are in bold. \label{table:seg_main_table}}
\begin{tabular}{l|c|c|c}
                                                  & \multicolumn{1}{c}{\textbf{\begin{tabular}[c]{@{}c@{}}IoU w/o \\ distractors\end{tabular}}} & \multicolumn{1}{c}{\textbf{\begin{tabular}[c]{@{}c@{}}IoU w\\  distractors\end{tabular}}} & \multicolumn{1}{c}{\textbf{Average}} \\ \hline
\multicolumn{1}{l|}{\textbf{FiLM}}                & \textbf{0.91}          & \textbf{0.84}                   & \textbf{0.86}            \\
\multicolumn{1}{l|}{\textbf{Concatenation}}       & 0.90             & 0.83                         & 0.85               \\
\multicolumn{1}{l|}{\textbf{Tiling}} &  0.90             & 0.82                                              & 0.84                         
\end{tabular}
\vspace{-0.4cm}
\end{table}

\subsection{Conditional Image Segmentation Evaluation} \label{subsec:cond-im-seg}
\subsubsection{\textbf{Motivation}}
In this section we aim to answer the following question : What is the best neural network architecture to use for image-conditioned object segmentation? Since this has to the best of our knowledge not been studied much in literature, in this work we benchmark three natural alternatives that can be applied to the problem.

\subsubsection{\textbf{Models Benchmarked}} The first model we built (\textit{Concatenation}) is a U-Net~\cite{unet} architecture, with its input being a channel-wise concatenation of the bottleneck and live images. The second model (\textit{Tiling}) first processes the bottleneck image through a convolutional network (CNN) to create a $256$-dimensional conditioning vector. This conditioning vector is then tiled and concatenated channel-wise to the feature maps at each resolution level of the encoder of a U-Net architecture that processes the live image. The third model we built is the FiLM-based architecture presented in Section~\ref{subsubsec:image-conditioned-seg}. In order to keep the comparison fair we keep as similar as possible both the U-Net architectures that forms the basis of all three methods, and the CNNs that process the bottleneck image in the Tiling and FiLM architectures. Full details of all three architectures can be found in our supplementary material.

\subsubsection{\textbf{Procedure}} In order to evaluate our three models, we created a test set with real-world images. We selected 13 random objects, and for each took 20 top-down images from our robot's camera at random positions and orientations, recreating the conditions present in our control evaluation. From these 20 images, 5 were without distractors. We then segmented by hand all of the 260 resulting images, giving us ground truth labels to compare against. We present the mean IoU between the network predictions and our ground truth labels for each architecture across the test images with and without distractors in Table~\ref{table:seg_main_table}.

\subsubsection{\textbf{Results}}

Interestingly, we see from Table~\ref{table:seg_main_table} that all three models performed similarly on the test set, with the FiLM architecture being slightly ahead.  During our experimentation,  we also noticed a much larger influence on the final performance from the training dataset's size and content, which we discuss in Section~\ref{subsec:seg_ablations}.

\begin{figure}[t!]   
\centering    
\includegraphics[width=0.45\textwidth]{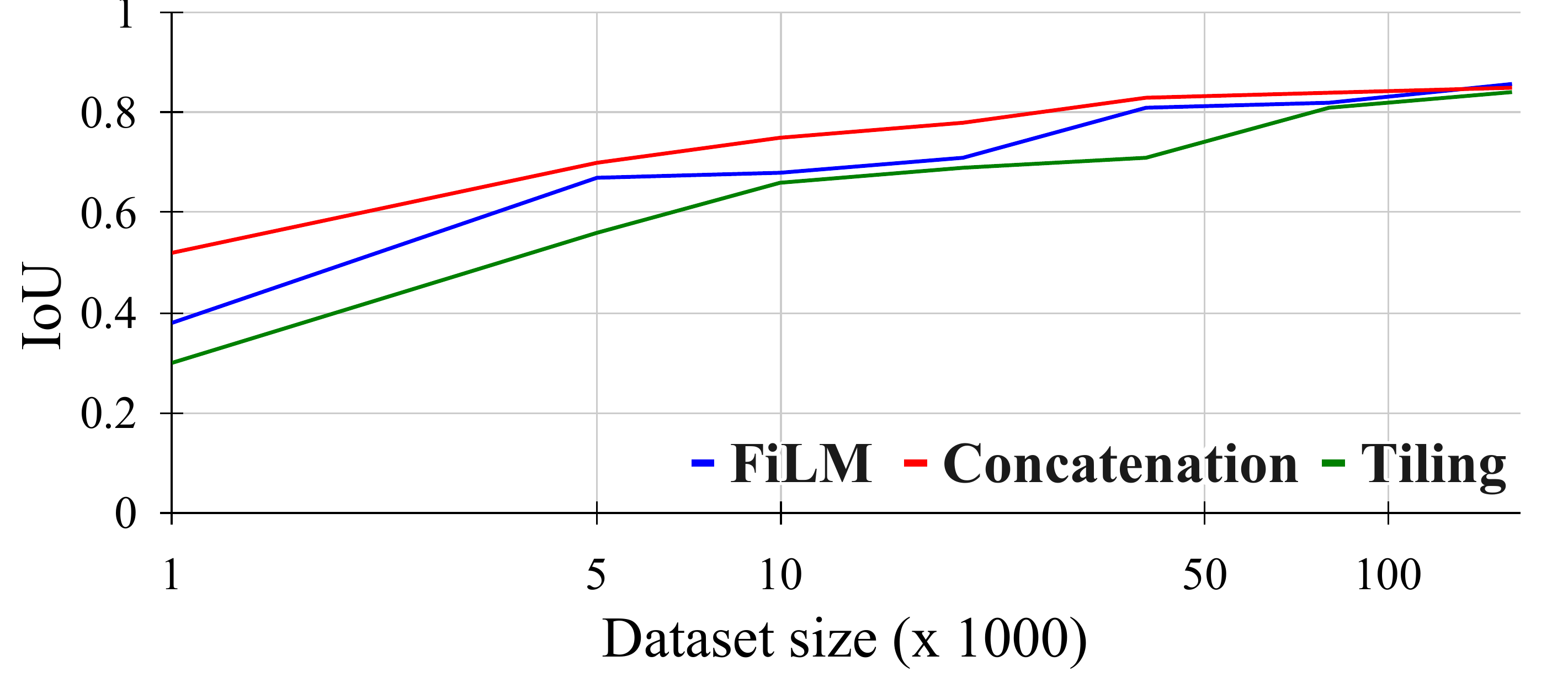}  
\caption{Segmentation network dataset size ablation. }
\label{fig:dataset_size}
\vspace{-0.4cm}
\end{figure}

\begin{figure}[tb]   
\centering    
\includegraphics[width=0.45\textwidth]{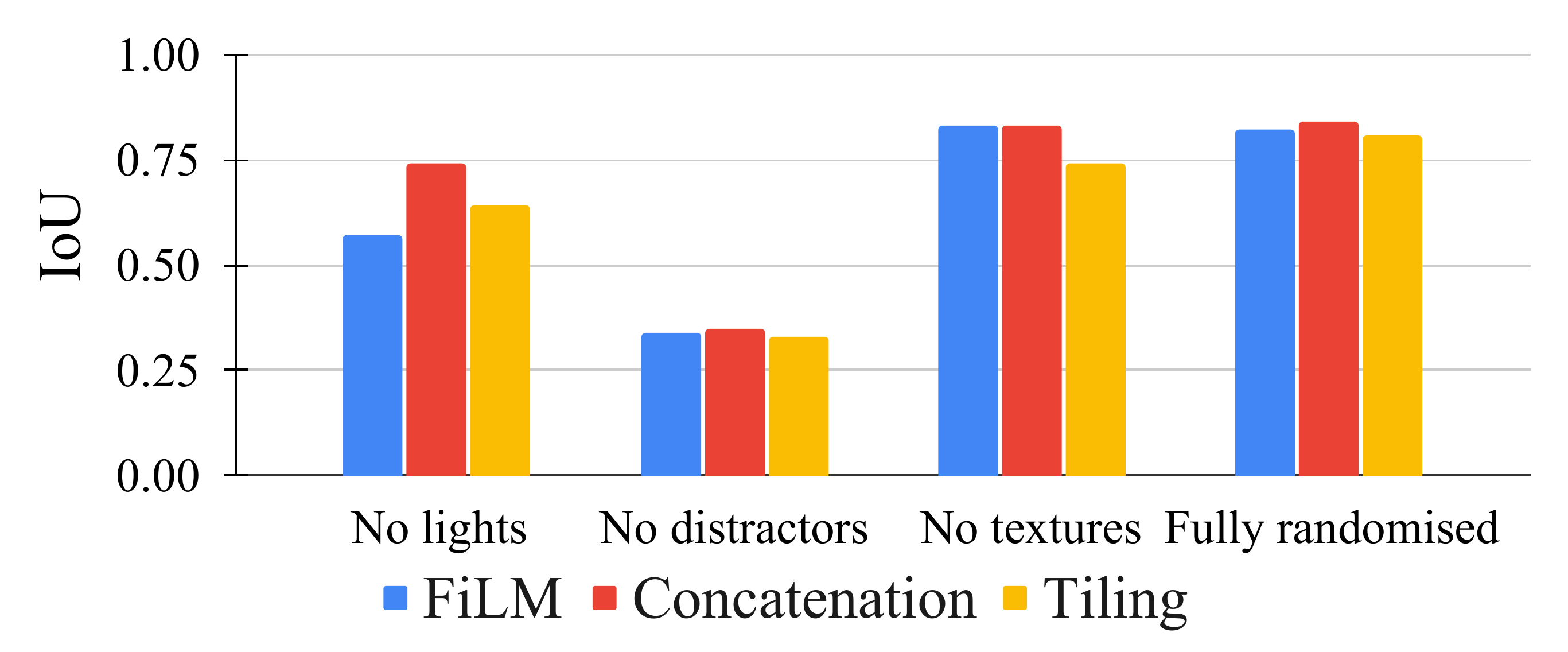}  
\caption{Segmentation network dataset content ablation. }
\label{fig:dr_ablation}
\vspace{-0.5cm}
\end{figure}

\subsection{Conditional Image Segmentation Ablations}\label{subsec:seg_ablations}
\subsubsection{\textbf{Motivation}}

In this section, we present two dataset ablations that we conducted for our conditional object segmentation models and that aim to answer the following questions: (1) How does their performance scale with the number of training data?, (2) How much are the different randomised aspects of the simulator responsible for their final performance?, and (3) Are any of the models tested more robust to some of these aspects than others?

\subsubsection{\textbf{Procedure}} In order to answer these questions, we first trained our three segmentation models on datasets of $\{1,5,10,20,40,80,160\}\times1000$ images, recorded their average IoU score across all our test images and plotted the results in Fig.~\ref{fig:dataset_size}. Second, we created three new $80,000$ image datasets, with the following differences:
\begin{itemize}
    \item \textit{No lights}, with no light source randomisation, using instead a single ambient light source,
    \item \textit{No textures}, without using random images as object textures, instead only randomising object properties discussed in Section~\ref{subsec:training-data},
    \item \textit{No distractors}, without any distractor objects.
\end{itemize}

We again trained all three models on each of these datasets and the fully randomised $80,000$ image dataset, and report their average IoU performance on the test set in Fig.~\ref{fig:dr_ablation}.

\subsubsection{\textbf{Results}}
Regarding the dataset size effect, we clearly see that as the dataset size increases the performance of all models also increases. Interestingly, we note that the Concatenation model performs best in all but the largest data size, where FiLM performed best (see Table~\ref{table:seg_main_table}).

For the domain randomisation ablations on the dataset, we first notice that having light source randomisation and distractor objects randomisation as part of the dataset generation are essential to achieving good final performance. Interestingly, we then notice that not assigning random images as textures to the objects has the least effect, thereby indicating that simply randomising object surface properties (such as colour or surface roughness) in a high fidelity simulator like blender may be sufficient for crossing the reality gap, which is also consistent with previous studies on the subject~\cite{denninger2020blenderproc}.

\subsection{Limitations and Future Work}
Despite the great capabilities shown by our method, we believe it is important to discuss some of its limitations and failure modes, which we aim to address in future work. First, our method does not account for having other objects appearing in the bottleneck image, or having the bottleneck object being partially out of frame, but we believe it may be possible to address this by adjusting the simulation training distributions. Second, the user has to carefully set the bottleneck pose to ensure the object is well within the camera's view, which should be improved to allow non-experts to provide demonstrations naturally. Finally, we cannot change the object to manipulate between demonstration and deployment. Indeed DOME's training does not account for this, and even small geometric changes in the object could (1) make the bottleneck pose become ill-defined in deployment, and (2) not allow simply replaying demonstration velocities to complete the task. Addressing these issues are exciting avenues for future work, and we believe are the correct path to a system able to achieve imitation learning in a truly practical and general way.

\section{Conclusions}

In this paper we have presented DOME, a method that can perform a task immediately following a single demonstration of that task. We have shown DOME's capabilities on 7 real-world everyday tasks, where it achieved near $100\%$ success and outperformed a number of baselines, even though these required further data collection and training beyond the single demonstration. To the best of our knowledge, this is the first imitation learning method that can be deployed immediately following a demonstration, without requiring any further data collection or training after the demonstration, or prior knowledge of the task or object. As such, DOME takes the field a step closer to the promise of practical one-shot imitation learning.

\addtolength{\textheight}{-12cm}   





\bibliography{references}
\bibliographystyle{abbrv}

\end{document}